\documentclass[11pt]{article}
\usepackage{coling2020}
\usepackage{times}
\usepackage{url}
\usepackage{latexsym}
\usepackage{graphicx}
\usepackage{multirow}
\usepackage{subfigure}
\usepackage{enumitem}
\usepackage{color, colortbl, xcolor}
\usepackage{bm}
\usepackage{amsmath}
\usepackage{amsfonts}
\usepackage{array}
\usepackage{url}
\usepackage{amsthm}
\usepackage{amssymb}

\newcolumntype{L}[1]{>{\raggedright\arraybackslash}p{#1}}
\newcolumntype{C}[1]{>{\centering\arraybackslash}p{#1}}
\newcolumntype{R}[1]{>{\raggedleft\arraybackslash}p{#1}}

\usepackage[colorlinks,
            urlcolor=magenta,
            linkcolor=red,
            anchorcolor=blue,
            citecolor=blue,
            ]{hyperref}

\colingfinalcopy % Uncomment this line for the final submission

\title{Exploring Question-Specific Rewards for Generating Deep Questions}

\author{Yuxi Xie$^{1}$ \quad Liangming Pan$^{2,4}$\thanks{\quad Corresponding author.} \quad Dongzhe Wang$^{3}$\\
\textbf{Min-Yen Kan$^4$ \quad Yansong Feng$^{1,5}$}\\
$^1$Wangxuan Institute of Computer Technology, Peking University\\
$^2$NUS Graduate School for Integrative Sciences and Engineering\\
$^3$ZhuiYi Technology, Singapore\\
$^4$School of Computing, National University of Singapore, Singapore\\
$^5$ The MOE Key Laboratory of Computational Linguistics, Peking University\\
{\tt xieyuxi@pku.edu.cn, e0272310@u.nus.edu, wdzethan2010@gmail.com} \\
{ \tt kanmy@comp.nus.edu.sg, fengyansong@pku.edu.cn} \\
}

\date{}

\begin{document}
\maketitle

\begin{abstract}
    Recent question generation (QG) approaches often utilize the sequence-to-sequence framework (Seq2Seq) to optimize the log likelihood of ground-truth questions using teacher forcing. However, this training objective is inconsistent with actual question quality, which is often reflected by certain global properties such as whether the question can be answered by the document. As such, we directly optimize for QG-specific objectives via reinforcement learning to improve question quality. We design three different rewards that target to improve the fluency, relevance, and answerability of generated questions. We conduct both automatic and human evaluations in addition to thorough analysis to explore the effect of each QG-specific reward. 
    We find that optimizing on question-specific rewards generally leads to better performance in automatic evaluation metrics. However, only the rewards  that correlate well with human judgement (e.g., relevance) lead to real improvement in question quality. Optimizing for the others, especially answerability, introduces incorrect bias to the model, resulting in poor question quality. Our code is publicly available at \href{https://github.com/YuxiXie/RL-for-Question-Generation}{https://github.com/YuxiXie/RL-for-Question-Generation}.
\end{abstract}

\section{Introduction}

% What is QG and Why it is important
Question Generation (QG) aims to endow machines with the ability to ask relevant and to-the-point questions about a document. 
QG has important practical applications, such as 
generating assessments for course materials in education~\cite{DBLP:conf/naacl/HeilmanS10,DBLP:conf/enlg/LindbergPNW13}, prompting user interaction in dialog systems~\cite{DBLP:conf/acl/ShuklaESKTW19}, enabling machines to ask clarification questions such as FAQs~\cite{DBLP:conf/emnlp/SaeidiBL0RSB018,DBLP:conf/acl/KrishnaI19}, and automatically building large-scale QA datasets for the research community~\cite{DBLP:conf/acl/DuSC17,DBLP:conf/emnlp/ZhaoNDK18}. 

% How tranditional works do it?
Recent QG approaches~\cite{DBLP:conf/acl/DuSC17,DBLP:conf/emnlp/ZhaoNDK18,DBLP:conf/www/LiuZNLHWX19} have used Seq2Seq models with attention~\cite{DBLP:journals/corr/BahdanauCB14}, which feeds the input document into an encoder, and generates a question about the document through a decoder. 
% Why it needs RL?
The training objective is to maximize the log likelihood of the ground-truth question paired with each input document using teacher forcing~\cite{DBLP:journals/neco/WilliamsZ89}. However, as the ground-truth questions are insufficient to account for the many equivalent ways of asking a question, this likelihood-based training suffers from the problem of exposure bias~\cite{DBLP:journals/corr/RanzatoCAZ15}, \textit{i.e.}, the model does not learn how to distribute probability mass over sequences that are valid but different from the ground truth. 
% How RL addresses the problem?
% (\textit{e.g.}, fluency, answerability) 
To address this issue, previous QG works proposed to optimize the model directly on \textit{question-specific rewards} via Reinforcement Learning (RL).  This process decouples the training procedure from the ground truth data, so that the space of possible questions can be better explored. Moreover, it allows the training to target on specific properties we want the question to exhibit, such as relevant to a specific topic or answerable by the document. 
% What is the problem for RL-based method?
Although various rewards have been employed for QG --- such as BLEU~\cite{Kumar2018PuttingTH}, the answerability reward~\cite{Zhang2019AddressingSD}, and the word movers distance~\cite{DBLP:conf/iclr/0022WZ20} --- optimizing the reward scores does not always lead to higher question quality in practice, as observed by Hosking and Riedel~\shortcite{Hosking2019EvaluatingRF}. How to define robust and effective QG-specific rewards still requires further investigation. 

% What we want to do?
We aim to analyze the effectiveness of question-specific rewards in QG. Instead of using general natural language generation metrics such as BLEU, we target three QG-related metrics that are commonly cited in human evaluations of question quality: (1) \textbf{Fluency} indicates whether the question follows the grammar and accords with the correct logic; (2) \textbf{Relevance} indicates whether the question is relevant to the document; and (3) \textbf{Answerability} indicates whether the question is answerable given the document. We design a specific RL reward for each metric: a language model based reward for fluency, a discriminator-based reward for relevance, and a QA-based reward for answerability. 
After optimizing each reward via RL, we conduct comprehensive analysis, including automatic and human evaluation, to arrive at the following conclusions: (1) both individual and joint optimization of these rewards can lead to performance gain in automated metrics, but this does not guarantee an improvement in the real question quality; (2) the reward for relevance substantially helps to improve the question quality, while the reward for answerability reduces the quality due to the bias brought by the QA model; and (3) a reward is more likely to improve the question quality if the reward score correlates well with human judgement. 

\section{Related Work}
% NQG related works
Early QG studies focused on using manually-designed rules or templates to transform a piece of given text to questions~\cite{heilman2011automatic,DBLP:conf/coling/ChaliH12a}, with low generalizability and scalability. To address this, recent neural question generation (NQG) models take advantage of the Seq2Seq framework with attention, which are trained in an end-to-end manner, requiring far less labor and enabling better language flexibility. Many improvements have been made to the original Seq2Seq NQG model~\cite{DBLP:conf/acl/DuSC17}, such as encoding answer information~\cite{DBLP:conf/nlpcc/ZhouYWTBZ17,DBLP:conf/emnlp/SunLLHMW18} and incorporating linguistic features~\cite{DBLP:conf/www/LiuZNLHWX19}. A comprehensive survey of QG can be found in~\cite{DBLP:journals/corr/abs-1905-08949}. 

% Existing Rewards
% replaced these $n$-gram based similarities with
Among these attempts, utilizing RL to optimize QG-specific rewards has been adopted by recent works to address the exposure bias problem. To find a good proxy for question quality, various rewards have been proposed. One common type of reward is the similarity between the generated question and the reference question written by human. Kumar \textit{et al.}~\shortcite{Kumar2018PuttingTH} adopted BLEU, ROUGE, and METEOR as rewards. Followup works employed more semantic-relevant metrics, such as the word movers distance~\cite{DBLP:conf/iclr/0022WZ20,yu_2020_low} and the paraphrasing probability~\cite{Zhang2019AddressingSD}. To generate more passage-relevant questions, Kumar \textit{et al.}~\shortcite{Kumar2018PuttingTH} designed a reward to measure the relevance between the input passage and the generated question based on their degree of overlapping. The \textit{answerability} reward measures whether the generated question can be answered by the input passage. It is designed as either the confidence score that a pre-trained QA model can correctly answer the generated question~\cite{Zhang2019AddressingSD}, or the overlapping degree between the target answer and the answer predicted by the QA model~\cite{yu_2020_low}. Other types of rewards include Yao \textit{et al.}~\shortcite{Yao2018TeachingMT}, which train a discriminator to measure the \textit{naturalness}, \textit{i.e.}, the question is human-written or generated. 

Most question-specific rewards are empirically successful since they achieve performance gain in automatic evaluation metrics after RL training. However, this brings several followup questions that existing works have failed to answer: (1) does optimizing RL rewards really improve the question quality from the human standard, (2) which reward is more effective in improving the question quality, and (3) how the rewards interfere with each other when jointly optimized. This paper aims to bridge this gap through human evaluation and analytic experiments, aiming to provide a better understanding of how different rewards affect the question generation process. 

\section{Methodology}
\label{sec:method}

\begin{figure}
    \centering
    \includegraphics[width=0.9\linewidth]{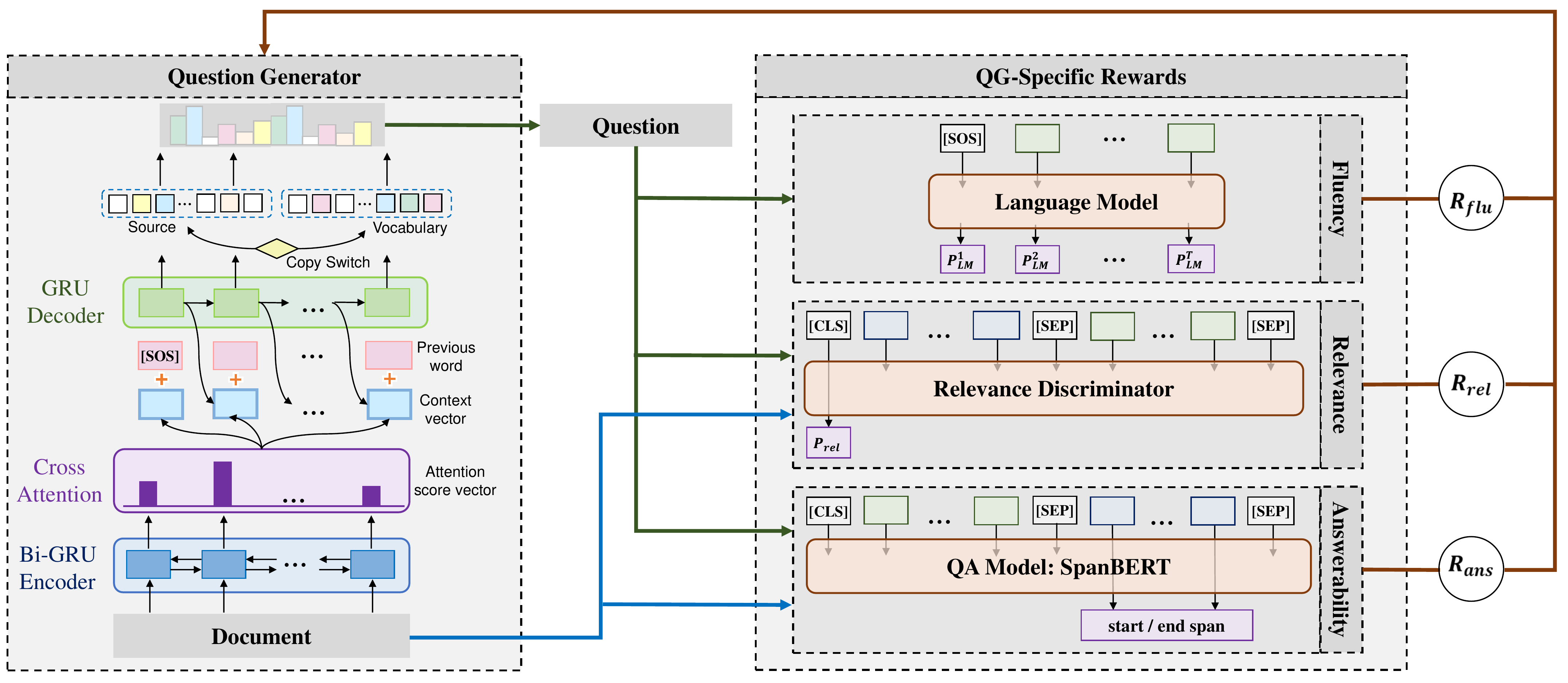}
    \caption{The framework of our model, consisting of the basic question generator (on the left) and the discriminators for QG-specific rewards (on the right). The blue sequence on the right represents the input document, and the green sequence is the generated question.}
    \label{fig:framework}
\end{figure}

Given a document $\mathcal{D}$ as input, the objective is to generate a relevant question $\hat{\mathcal{Y}}$ which can be answered by the document $\mathcal{D}$. This is formulated as maximizing the conditional probability $p(\mathcal{Y}|\mathcal{D})$: 
\begin{equation}
    \label{equ:general}
    \hat{\mathcal{Y}} = \arg\max_{\mathcal{Y}}P(\mathcal{Y}|\mathcal{D})=\arg\max_{\mathcal{Y}}\prod^T_{t=1}P(y_t|\mathcal{D},\mathcal{Y}_{<t})
\end{equation}
where $y_t$ is the $t$-th token of the generated question $\mathcal{Y}$, and $\mathcal{Y}_{<t}$ represents the previous decoded tokens, \textit{i.e.}, $y_1, \cdots, y_{t-1}$. The general framework of our model is shown in Figure~\ref{fig:framework}, consisting of two parts: the \textit{Question Generator} and the \textit{QG-specific Rewards}. The \textbf{Question Generator} uses the Seq2Seq framework with attention~\cite{DBLP:journals/corr/BahdanauCB14}, copying~\cite{DBLP:conf/acl/GuLLL16,DBLP:conf/acl/SeeLM17}, and coverage mechanisms~\cite{DBLP:conf/acl/TuLLLL16}, following most existing NQG works. The model is trained by maximizing the likelihood of ground-truth questions. As discussed in the introduction, this basic question generator suffers from the exposure bias problem. Therefore, we design three \textbf{QG-Specific Rewards} aiming at evaluating the fluency, relevance, and answerablity of the question generated by the basic model. We then fine-tune the model by optimizing these rewards following the RL framework with a baseline~\cite{Rennie_2017}. In the following, we describe the design of the three QG-specific rewards in detail. 

\subsection{LM-based Reward for Fluency}
\label{subsec:fluency}

The perplexity of a sentence under a well-trained Language Model (LM) usually serves as a good indicator of its fluency~\cite{DBLP:conf/nips/YangHDXB18}. Therefore, we introduce an LM-based reward to improve the fluency of the generated question. We first pre-train a language model $P_{LM}$ and then define the \textbf{fluency reward} $\mathcal{R}_{flu}$ of a generated question $\mathcal{Y}$ as its negative perplexity evaluated by $P_{LM}$, formulated as: 
\begin{equation}
    \label{equ:fluency-reward}
    \mathcal{R}_{flu}(\mathcal{Y}) = - \exp(-\frac{1}{T}\sum_{t=1}^T\log{P_{LM}(y_t|\mathcal{Y}_{<t})})
\end{equation}
To optimize the fluency reward in training, we define the following loss function $\mathcal{L}_{flu}$: 
\begin{equation}
    \label{equ:fluency-loss}
    \mathcal{L}_{flu} = -(\mathcal{R}_{flu}(\hat{\mathcal{Y}}) - \alpha_{flu})\frac{1}{T}\sum_{t=1}^T\log{P_{QG}(\hat{y}_t|\mathcal{D}, \hat{\mathcal{Y}}_{<t})}
\end{equation}
where $\hat{y}_t$ is the $t$-th token in the predicted question $\hat{\mathcal{Y}}$, which is sampled from the vocabulary distribution $P_{QG}(y_t|\mathcal{D}, Y_{<t})$ specified by the RNN decoder of the question generator. $\alpha_{flu}$ is a pre-defined negative perplexity, which is used as the baseline reward in the RL algorithm to stabilize the training process. 

\subsection{Discriminator-Based Reward for Relevance}

We then design a classifier-based discriminator to judge whether the generated question is relevant to the input document. As shown in Figure~\ref{fig:framework}, the discriminator is a binary classifier based on the pre-trained BERT~\cite{DBLP:conf/naacl/DevlinCLT19}, which takes both the input document $\mathcal{D}$ and the generated question $\mathcal{Y}$ as inputs and outputs the probability that $\mathcal{Y}$ is relevant to the $\mathcal{D}$. To train the relevance discriminator, we use the human-written ground-truth questions $\mathcal{Y}_G$ for each document as the positive training data. For a document-question pair $(\mathcal{D}, \mathcal{Y}_G)$, we create the negative sample $\mathcal{Y}_N$ for $\mathcal{D}$ in the following three ways:

% \paragraph
\textbf{$\bullet$ Question Swap} We randomly select a ground-truth question from another document $\mathcal{D}'$ as the negative sample for the document $\mathcal{D}$. 

\textbf{$\bullet$ Inter-Doc Entity Swap} We create the negative sample $\mathcal{Y}_N$ by replacing the entity in the ground-truth question $\mathcal{Y}_G$ with another entity of the same type but does not occur in the document $\mathcal{D}$. This helps the discriminator to learn whether the question involves entities not mentioned in the document. 

\textbf{$\bullet$ Intra-Doc Entity Swap} We also replace the entity in the ground-truth question with a different entity from the same document. This often creates logical errors in the question, \textit{e.g.}, \textit{William Shakespeare is written by the book}, which is more challenging for the discriminator to differentiate. 

Following the above process, we create three negative samples for each ground-truth question. To address the unbalance between positive and negative training data, we adopt the $\alpha$-balanced focal loss~\cite{Lin_2017} to train the relevance discriminator, given as follows. 
\begin{equation}
    \label{equ:relevance-train}
    \mathcal{L}_F(P_t) = -\alpha_t(1-P_t)^{\lambda}\log{P_t}
\end{equation}
where $P_t$ is the predicted probability for class $t$. $(1-P_t)^\lambda$ is a modulating factor with a tunable focusing parameter $\lambda \geq 0$ that smoothly adjusts the rate at which easy examples are down-weighted. 

After training the relevance discriminator, we use it to obtain the relevance reward and then fine-tune the question generator by maximizing the relevance reward in RL training. Given a document $\mathcal{D}$ and a question $\mathcal{Y}$, the \textbf{relevance reward} $\mathcal{R}_{rel}(\mathcal{D}, \mathcal{Y})$ is defined as a scaling of the relevance probability $P_{rel}(\mathcal{D}, \mathcal{Y})$ output by the relevance discriminator as: 
\begin{equation}
    \label{equ:relevance-reward}
    \mathcal{R}_{rel}(\mathcal{D}, \mathcal{Y}) = -\log{(1 - P_{rel}(\mathcal{D}, \mathcal{Y}) + \epsilon)}
\end{equation}
where $\epsilon$ is a positive factor close to zero to avoid calculating $\log{0}$. We scale the relevance probability in this way to augment the reward value for positive samples, \textit{i.e.,} those samples whose rewards are greater than the baseline, because the QG model generally samples more negative samples during training. To optimize the relevance reward in RL training, we define the loss function $\mathcal{L}_{rel}$ as follows. 
\begin{equation}
    \label{equ:relevance-loss}
    \mathcal{L}_{rel} = -(\mathcal{R}_{rel}(\mathcal{D}, \hat{\mathcal{Y}}) - \alpha_{rel})\frac{1}{T}\sum_{t=1}^T\log{P_{QG}(\hat{y}_t|\mathcal{D}, \hat{\mathcal{Y}}_{<t})}
\end{equation}

\subsection{QA-Based Reward for Answerability}

Answerability indicates whether the question is answerable by the document without the need of external information. We design the \textbf{answerability reward} based on the SpanBERT~\cite{DBLP:journals/tacl/JoshiCLWZL20}, a state-of-the-art model for extractive QA. Given a document $\mathcal{D}$ and a question $\mathcal{Y}$ as inputs, SpanBERT predicts the start and end spans of the potential answer in the document $\mathcal{D}$. Formally, it outputs two probability distributions over the tokens in the document: $P_{ans}^s$ and $P_{ans}^e$, where $P_{ans}^s(i) / P_{ans}^e(i)$ is the probability that the $i$-th token is the start/end span of the answer. Based on the pre-trained SpanBERT model, we first fine-tune it with the HotpotQA dataset~\cite{DBLP:conf/emnlp/Yang0ZBCSM18} and then use it to obtain the answerability reward for the generated question $\mathcal{Y}$. Intuitively, when the question is answerable, the model should be quite confident about the start/end span of the answer, so the distribution should be peak for both $P_{ans}^s$ and $P_{ans}^e$, \textit{i.e.}, the value of $\max_i P_{ans}^s (i)$ and $\max_j P_{ans}^e (j)$ are both large. Therefore, we use the geometric average of these two values to indicate the answerability, formulated as follows. 
\begin{equation}
    \label{equ:answerability-reward}
    \mathcal{R}_{ans}(\mathcal{D}, \mathcal{Y}) = -\log{(1-\max_{1\leq i \leq j \leq T, \ j-i\leq l}\sqrt{P_{ans}^s (i \vert \mathcal{D}, \mathcal{Y}) \cdot P_{ans}^e(j \vert \mathcal{D}, \mathcal{Y})} + \epsilon)}
\end{equation}
where $l$ represents the maximum allowed length of the answer. Similar to Equation~\ref{equ:relevance-reward}, we also scale the probability to balance positive and negative samples during training. Similar to previous sections, to optimize the answerability reward in RL training, we define a loss function $\mathcal{L}_{ans}$: 
\begin{equation}
    \label{equ:answerability-loss}
    \mathcal{L}_{ans} = -(\mathcal{R}_{ans}(\mathcal{D}, \hat{\mathcal{Y}}) - \alpha_{ans})\frac{1}{T}\sum_{t=1}^T\log{P_{QG}(\hat{y}_t|\mathcal{D}, \hat{\mathcal{Y}}_{<t})}
\end{equation}

\subsection{Model Training}

We train the whole model following the pre-training and fine-tuning paradigm, as in~\cite{Hosking2019EvaluatingRF}. We first pre-train the question generation model by minimizing the cross-entropy loss together with the copying loss and the coverage loss, which can be written together as $\mathcal{L}_{base}$:
\begin{equation}
    \label{equ:base-loss}
    \mathcal{L}_{base} = \frac{1}{T}\sum_{t=1}^T(-\log{P_{QG}(\hat{y}_t|\mathcal{D}, \mathcal{Y}}_{<t})+\gamma_{cov}\sum_i\min(a_i^t, c_i^t))
\end{equation}
where the copy mechanism is involved in the question generator $P_{QG}$, and $a_i^t$ is the $i^{th}$ element of the attention score vector over the document at time stamp $t$. We then fine-tune the basic QG model trained with $\mathcal{L}_{base}$ to maximize the previously defined QG-specific rewards. This is achieved by linearly combining $\mathcal{L}_{base}$ with the RL-based losses $\mathcal{L}_{flu}, \mathcal{L}_{rel}, \mathcal{L}_{ans}$, as follows. 
\begin{equation}
    \label{equ:loss}
    \mathcal{L} = \mathcal{L}_{base} + \gamma_{flu}\mathcal{L}_{flu} + \gamma_{rel}\mathcal{L}_{rel} + \gamma_{ans}\mathcal{L}_{ans}
\end{equation}
where the hyper-parameters $\gamma_{flu}$, $\gamma_{rel}$ and $\gamma_{ans}$ specify the trade-off between different kinds of rewards. Note that we empirically set baseline rewards $\alpha_{flu}$, $\alpha_{rel}$, and $\alpha_{ans}$ to reduce the variance of gradient estimation during RL training, as reflected in Equations~\ref{equ:fluency-loss}, \ref{equ:relevance-loss}, and \ref{equ:answerability-loss}.  

\section{Experiments}

% dataset & why we choose it
We conduct experiments on HotpotQA~\cite{DBLP:conf/emnlp/Yang0ZBCSM18}, containing $\sim$100K crowd-sourced questions paired with Wikipedia articles. Generating a fluent, relevant, and answerable question in HotpotQA is a non-trivial task as it requires reasoning over different pieces of information in the input document. 
% data split
We follow the data split of Pan \textit{et al.}~\shortcite{DBLP:conf/acl/PanXFCK20} to get 90,440 / 6,072 examples for training and testing, respectively. We further hold out 6,072 examples from the training data as the development set. 

% QG framework
The basic question generator is a Seq2Seq framework with  copying~\cite{DBLP:conf/acl/GuLLL16}, coverage~\cite{DBLP:conf/acl/SeeLM17}, and attention~\cite{Hou2019CrossAN} mechanisms. 
We employ a $1$-layer bi-directional GRU as the encoder and a $1$-layer GRU as the decoder. 
We use the cased WordPiece tokenizer for the question generator following Joshi \textit{et al.}~\shortcite{DBLP:journals/tacl/JoshiCLWZL20}. 
The hidden size of the Seq2Seq model and the maximal input sequence length are set as $512$ and $256$, respectively. 

To train the language model used for evaluating the fluency reward, we fine-tune the pre-trained BERT model~\cite{DBLP:conf/naacl/DevlinCLT19} on our target dataset, resulting in an LM with a perplexity of $8.85$ on the dev set. 
Our relevance discriminator, which is also fine-tuned from the pre-trained BERT model, achieves a $91.16$ $F_1$ score. The answerability discriminator based on SpanBERT-large obtains a $70.60$ Exact Match (EM) score and an $83.44$ F1 score on the HotpotQA development set. 
In RL training, we empirically set the baseline rewards $\alpha_{flu}$, $\alpha_{rel}$ and $\alpha_{ans}$ as $-10$, $\log(2)$, and $\log(2)$, respectively. When jointly training all the rewards, the trade-off parameters $\gamma_{cov}$, $\gamma_{flu}$, $\gamma_{rel}$ and $\gamma_{ans}$ are tuned on the dev set and set to $0.25$, $0.2$, $1$, $1$, respectively. We provide ancillary material about supplementary experiments on hyper-parameter sensitivity and result analysis in our codebase\footnote{\href{https://github.com/YuxiXie/RL-for-Question-Generation/tree/main/doc}{https://github.com/YuxiXie/RL-for-Question-Generation/tree/main/doc}}. Other settings for training follow the standard best practice\footnote{All models are trained using Adam~\cite{DBLP:journals/corr/KingmaB14} with mini-batch size $64$. The learning rate is initially set to $10^{-3}$, and adaptive learning rate decay applied. We also adopt early stopping and use gradient clipping~\cite{DBLP:conf/icml/PascanuMB13} with clip norm of $5$.}. 

\subsection{Automatic Evaluation}

\begin{table*}[t]
    \small
	\begin{center}
		\begin{tabular}{ l || c | c | c || c | c | c | c || c | c | c } \hline
        \multicolumn{1}{c||}{\multirow{2}{*}{\textbf{Model}}} & \multicolumn{3}{c||}{\textbf{Rewards}} & \multicolumn{7}{c}{\textbf{Metrics}} \\ 
        & \textbf{F} & \textbf{R} & \textbf{A} & \textbf{BLEU1} & \textbf{BLEU4} & \textbf{METEOR} & \textbf{ROUGE-L} & \textbf{R-FLU} & \textbf{R-REL} & \textbf{R-ANS} \\ \hline \hline
        B1. Baseline &  &  &  & 33.68 & 13.46 & \textbf{21.39} & 35.06 & $-$ & $-$ & $-$ \\ \hline % -6.00 & 3.20 & 1.36 \\ \hline
        S1. F & $\surd$ &  &  & 37.59$^*$ & 15.22$^*$ & 19.49$^*$ & 35.08 & +1.48$^*$ & +0.49$^*$ & +0.03 \\ \hline
        S2. R &  & $\surd$ &  & 36.33$^*$ & 14.83$^*$ & 20.63$^*$ & \textbf{35.58$^*$} & +1.06$^*$ & \textbf{+0.61$^*$} & +0.04 \\ \hline
        S3. A &  &  & $\surd$ & 36.40$^*$ & 13.95$^*$ & 18.73$^*$ & 34.07$^*$ & +1.30 & +0.18$^*$ & +0.21$^*$ \\ \hline
        E1. F + R & $\surd$ & $\surd$ &  & 37.82$^*$ & 15.30$^*$ & 19.95$^*$ & 35.48$^*$ & +1.30$^*$ & +0.60$^*$ & +0.03 \\ \hline
        E2. R + A &  & $\surd$ & $\surd$ & 35.77$^*$ & 14.46$^*$ & 20.53$^*$ & 35.26 & +0.78 & +0.49$^*$ & +0.36$^*$ \\ \hline
        E3. F + A & $\surd$ &  & $\surd$ & \textbf{38.30$^*$} & 14.99$^*$ & 18.02$^*$ & 34.50$^*$ & \textbf{+1.71$^*$} & +0.40$^*$ & \textbf{+0.51$^*$} \\ \hline
        E4. F + R + A & $\surd$ & $\surd$ & $\surd$ & 37.97$^*$ & \textbf{15.41$^*$} & 19.61$^*$ & 35.12 & +1.57$^*$ & \textbf{+0.61$^*$} & +0.49$^*$ \\ \hline
		\end{tabular}
	\end{center}
	% MinCR: actually you don't need columns 2-4 since your name already connotes the additions.  But fine to leave them.  For next time.
\caption{The QG performance evaluated by automatic metrics when separately or jointly optimizing for various rewards. The last three columns show the change of reward scores compared with B1, where \textbf{R-FLU} is the fluency, \textbf{R-REL} the relevance, and \textbf{R-ANS} the answerability rewards. $*$ denotes that the increase/drop in performance compared with B1 is statistically significant for $p < 0.01$.}
\label{tbl:reward_auto_performance}
\end{table*}

To investigate the effect of different QG-specific rewards, we first report the performance of automatic evaluation metrics when optimizing different rewards. The metrics include: a) BLEU~1 and 4~\cite{DBLP:conf/acl/PapineniRWZ02}, METEOR~\cite{DBLP:conf/wmt/LavieA07}, and ROUGE-L~\cite{lin-2004-rouge}, which are based on the $n$-gram similarity between the generated questions and the ground truth; and b) gain on reward scores (the fluency, relevance, answerability rewards) after RL training. Table~\ref{tbl:reward_auto_performance} summarizes the performance, where B1 is the basic question generator without RL training. The other models are fine-tuned based on B1 by optimizing either a single (S1--S3) or multiple rewards together (E1--E4). F, R, and A represents the fluency, relevance reward, and answerability rewards, respectively. We make four major observations:

1. Optimizing a single reward alone (F, R, A) can lead to an improvement on the BLEU score and also its corresponding reward score (F$\rightarrow$R-FLU, R$\rightarrow$R-REL, and A$\rightarrow$R-ANS). When optimizing one reward, the scores for the other two also slightly increase, showing that the three rewards are correlated. This is in line with our intuition; \textit{e.g.}, a question answerable by the passage is also likely to be fluent. 

2. Jointly training multiple rewards in general leads to better performance. For example, the best improvement of R-REL, R-FLU and R-ANS are achieved by E3 and E4. This shows that different rewards can mutually enhance each other in joint training, which provides a prospective future direction on RL reward integration.

3. In general, the increase in rewards do not correlate well with improvement on automatic metrics. For example, E3 has the largest reward gain in fluency and answerability, but achieves relatively worse results in BLEU4, METEOR and ROUGE-L. This shows that the RL rewards focus on different parts of the question quality other than the $n$-gram based similarity with the ground truth. We further investigate how each reward affects the question quality later in Section~\ref{sec:human-evaluation}. 

4. We find that our B1 baseline tends to generate longer questions (the average question length is $1.44$ times that of the ground truth, compared with $1.13$ for E4). The RL rewards thus encourage shortening to lengths which are closer to the ground truth. This explains why the improvements brought by RL rewards are especially significant on BLEU.

\subsection{Comparison with Baselines}

We then compare our best performing model (E4. F + R + A) against several strong baselines in QG. The technologies employed by each model as well as the performance results are summarized in Table~\ref{tbl:performance_comparision}. Without using the answer information and any external linguistic knowledge, our model achieves a comparative BLEU4 with the state-of-the-art QG model (B7) in HotpotQA. This demonstrates the effectiveness of optimizing QG-specific rewards via RL. Surprisingly, the CGC-QG (B6) model exhibits an unusual pattern, achieving the best METEOR and ROUGE-L, but worst BLEU1 among all baselines. 
% MinCR: do you have examples to show for this?  If not in the paper, in the github repo ancillary materials?
Our analysis finds that CGC-QG tends to generate irrelevant word during word-level content selection, leading to lengthy questions that are unanswerable or which contain semantic errors~\cite{DBLP:conf/acl/PanXFCK20}. 

\begin{table*}[!t]
    \small
	\begin{center}
		\begin{tabular}{ l || c | c | c | c | c | c || c | c | c | c } \hline
        \multicolumn{1}{c||}{\multirow{2}{*}{\textbf{Model}}} & \multicolumn{6}{c||}{\textbf{Features}} & \multicolumn{4}{c}{\textbf{Metrics}} \\
        & \textbf{AE} & \textbf{LF} & \textbf{CP} & \textbf{CV} & \textbf{SA} & \textbf{RL} & \textbf{BLEU1} &  \textbf{BLEU4} & \textbf{Meteor} & \textbf{Rouge-L} \\ \hline \hline
        B2. Bahdanau et al.~\shortcite{DBLP:journals/corr/BahdanauCB14} &  & &  &  &  &  & 32.97 & 11.81 & 18.19 & 33.48 \\ \hline
        B3. NQG++~\cite{DBLP:conf/nlpcc/ZhouYWTBZ17} &  & $\bullet$ & $\bullet$ &  & &  & 35.31 & 11.50 & 16.96 & 32.01 \\ \hline
        B4. Zhao et al.~\shortcite{DBLP:conf/emnlp/ZhaoNDK18} &  &  & $\bullet$ &  & $\bullet$ &  & 35.36 & 11.85 & 17.63 & 33.02 \\ \hline
        B5. Zhao et al.~\shortcite{DBLP:conf/emnlp/ZhaoNDK18} + ans, cov & $\bullet$ &  & $\bullet$ & $\bullet$ & $\bullet$ &  & 38.74 & 13.48 & 18.39 & 34.51 \\ \hline
        B6. CGC-QG~\cite{DBLP:conf/www/LiuZNLHWX19} & $\bullet$ & $\bullet$ & $\bullet$ & &  &  & 31.18 & 14.36 & \textbf{25.20} & \textbf{40.94} \\ \hline
        B7. SG-DQG~\cite{DBLP:conf/acl/PanXFCK20} & $\bullet$ & $\bullet$ & $\bullet$ & $\bullet$ & &  & \textbf{40.55} & \textbf{15.53} & 20.15 & 36.94 \\ \hline
        E4. Ours (F + R + A) &  &  & $\bullet$ & $\bullet$ & & $\bullet$ & 37.97 & 15.41& 19.61 & 35.12 \\ \hline
		\end{tabular}
	\end{center}
    \caption{Performance comparison. For all baselines, we use the reported performance from Pan \textit{et al.} (2020). Legend: \textbf{AE}: answer encoding, \textbf{LF}: linguistic features, \textbf{CP}: copying mechanism, \textbf{CV}: coverage mechanism, \textbf{SA}: gated self-attention, \textbf{RL}: reinforcement learning. }
    \label{tbl:performance_comparision}
\end{table*}

\subsection{Human Evaluation}
\label{sec:human-evaluation}

\begin{table*}[t]
    \small
	\begin{minipage}{0.42\textwidth}
	    \begin{center}
		\begin{tabular}{l|c|c|c|c} \hline
			\multirow{2}{*}{\textbf{Model}} & \textbf{Flu.} & \textbf{Rel.} & \textbf{Ans.} & \textbf{Cpx.} \\
			 & (1-5) & (1-3) & (0-1) & (1-3) \\ \hline
			B1. Baseline & 3.98 & 2.77 & 0.67 & \textbf{1.59} \\ \hline
			S1. F & 4.07 & 2.78 & 0.61 & 1.50 \\ \hline
			S2. R & \textbf{4.24} & \textbf{2.83} & \textbf{0.70} & 1.51 \\ \hline
			S3. A & 3.82 & 2.63 & 0.46 & 1.55 \\ \hline
			E4. F+R+A & 4.10 & 2.72 & 0.53 & 1.52 \\ \hline
		\end{tabular}
		\end{center}
		\caption{Human evaluation results for different methods. \textbf{Flu.}, \textbf{Rel.}, \textbf{Ans.}, and \textbf{Cpx.} denote the \textit{Fluency}, \textit{Relevance}, \textit{Answerability}, and \textit{Complexity}, respectively. }
		\label{tbl:human_evaluation}
	\end{minipage}
	\quad
	\begin{minipage}{0.55 \textwidth}
	    \begin{center}
	    \begin{tabular}{l} \hline
	         \textbf{Question with Options} \\ \hline
	         \textbf{Q1}. Whether this is a readable / understandable question? \\
	         $\circ$ yes \quad $\circ$ no \\ \hline
	         \textbf{Q2}. Which of the following errors occur in the question? \\ 
	         $\square$ correct \quad $\square$ repetition  $\square$ incomplete  $\square$ ambiguous reference \\
	         $\square$ incoherent \quad $\square$ wrong semantic collocation \quad $\square$ others \\ \hline
	         \textbf{Q3}. Whether this question is answerable by the passage?  \\ 
	         $\circ$ yes \quad $\circ$ no \\ \hline
	         \textbf{Q4}. Why the question is not answerable? \\ 
	         $\square$ invalid question \quad $\square$ ghost entity \\
	         $\square$ no ghost entity, but information insufficient \quad $\square$ others \\ \hline
	         \textbf{Q5}. Whether this question require reasoning to answer?  \\ 
	         $\circ$ yes, and very hard \quad $\circ$ yes, but simple reasoning \quad $\circ$ no \\ \hline
	    \end{tabular}
	    \end{center}
	    \caption{Questionnaire designed for human evaluation, where $\circ$ and $\square$ indicate single-item and multiple-item selection, respectively.}
	    \label{tbl:human_eval_questions}
	\end{minipage}
\end{table*}

To further investigate whether optimizing QG-specific rewards lead to real improvement in question quality, we conduct human evaluation on the generated questions for 200 randomly-sampled testing documents. 
We ask 6 workers to rate the questions generated by 5 different models: the basic question generator (B1), the models fine-tuned with a single reward (S1, S2, S3), and the model with all three rewards (E4).
Raters were blinded from the identity of the models.
We designed the scale differently for each metric to ease human rating effort.
For each question, we ask three workers to give ratings on four criteria: \textit{Fluency} (on a scale of 1--5), \textit{Relevance} (scaled 1--3), \textit{Answerability} (0 for unanswerable and 1 for answerable), and \textit{Complexity} (scaled 1--3). 
To reduce the subjectivity of human rating, we obtain the rating score according to the annotator's answers to our designed questionnaire shown in Table~\ref{tbl:human_eval_questions}. For more accurate evaluation, we give an unreadable question labeled by Q1 the lowest fluency rating and do not consider its relevance, answerability, and complexity ratings, as it is infeasible to judge them when the question is unreadable. The proportion of the unreadable questions generated by B1, S2, S2, S3 and E4 are $11.8\%$, $10.7\%$, $4.4\%$, $11.0\%$ and $10.0\%$, respectively. For a readable question, the fluency rating is determined by the number of grammar errors it makes (the answer to Q2). The answerability and complexity ratings are given by the answers of Q3 and Q5, respectively. The relevance score depends on both Q3 and Q4.
We average the scores from raters on each question, reporting the performance in Table~\ref{tbl:human_evaluation}. 
We discuss four major findings: 

1. Human ratings do not correlate well with automatic evaluation metrics (BLEU4, Meteor, ROUGE-L), showing that the $n$-gram based metric is not a good reflection of actual question quality. Similar observations also exist in other language generation tasks~\cite{DBLP:conf/eacl/Callison-BurchOK06,DBLP:conf/emnlp/LiuLSNCP16,DBLP:conf/emnlp/NovikovaDCR17} for fluency, adequacy and coherence, validating our findings.

2. Optimizing the relevance reward (S2) alone leads to a substantial improvement of the human ratings for fluency, relevance, and answerability. Our further analysis in Section~\ref{subsec:meso_analysis} shows that optimizing the relevance reward reduces ghost entity errors, a major source of error in previous QG models. 

3. In contrast, optimizing for answerability (S3) has a surprising negative effect: reducing scores against all three human ratings, compared against the baseline (B1). We believe this is due to the immature of the current QA model in answering deep questions; \textit{i.e.}, when used as a discriminator, the QA model we used cannot accurately predict whether a question is answerable or not, especially when the question involves reasoning (the case in HotpotQA). We analyse this in more depth in Section~\ref{subsec:violin_plot}. We also show in Section~\ref{subsec:meso_analysis} that the model tends to learn spurious correlations for answerability (\textit{e.g.}, a \textit{what year} question is more likely to be answerable), which also accounts for its poor performance.

4. The average length of the generated questions are $1.44$, $1.14$, $1.25$, $1.17$, and $1.13$ times compared with the ground-truth questions for B1, S1, S2, S3 and E4, respectively. This correlates with the human ratings for complexity ($1.59$, $1.50$, $1.51$, $1.55$, and $1.52$ for B1, S1, S2, S3 and E4 respectively). Longer questions tend to get higher complexity scores, which we feel validates intuition. 

\subsection{Consistency between Rewards and Human Judgement}
\label{subsec:violin_plot}

\begin{figure*}[!t]
    \small
    \centering
    \includegraphics[width=0.82\linewidth]{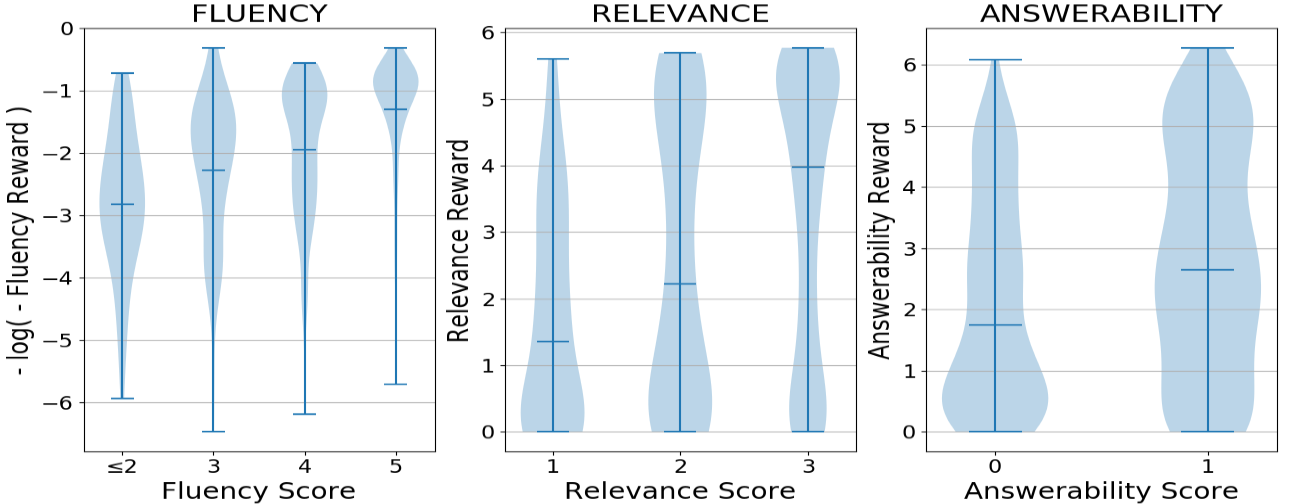}
    \caption{Correlation between reward scores and human ratings. Each sub-figure shows the distribution of reward scores on each level of human rating, where the hyphens in each column represent the minimal, median, and maximal values of the reward score.}
    \label{fig:violin-plot}
\end{figure*}

To figure out why certain rewards improve the real question quality while others do not, we plot violins to show the distribution of reward scores on each level of human rating, shown in Figure \ref{fig:violin-plot}. 

We observe that the relevance reward has the highest consistency with human judgement; \textit{i.e.}, both the median and the maximal rewards improves when the human rating gets higher. This provides an explanation of why optimizing the relevance reward leads to the best question quality. 

The answerability reward predicted by the QA model, however, exhibits a poor correlation with the true answerability judged by humans. The median answerability reward is low for both answerable and unanswerable questions labeled by humans. This lends evidence for our claim in Section~\ref{sec:human-evaluation} that the innate capability of the QA model is the bottleneck for this reward. We expect the answerability reward to become more effective as deep QA improves, and could become a key component in future work. 

The correlation between the fluency reward is also unsatisfactory: the increase of the fluency reward score is not obvious when the human rating for fluency increases.  This makes the performance of S1 similar to the baseline model in Table~\ref{tbl:human_evaluation}; \textit{i.e.}, the effect of optimizing the fluency reward is not obvious. 

Based on the above observations, we conclude that the rewards that correlate well with human judgements tend to achieve real improvement in question quality. Therefore, to design an effective QG-specific reward, testing its performance on $n$-gram based metrics such as BLEU may not faithfully reflect its effectiveness. Instead, running an initial test of how well the reward score correlates with human judgement seems more viable. 

\begin{figure*}
    \small
    \centering
    \subfigure[Correlation between human ratings]
    {
    	\begin{minipage}[t]{0.4\linewidth}
    	\centering
    	\includegraphics[width=\linewidth]{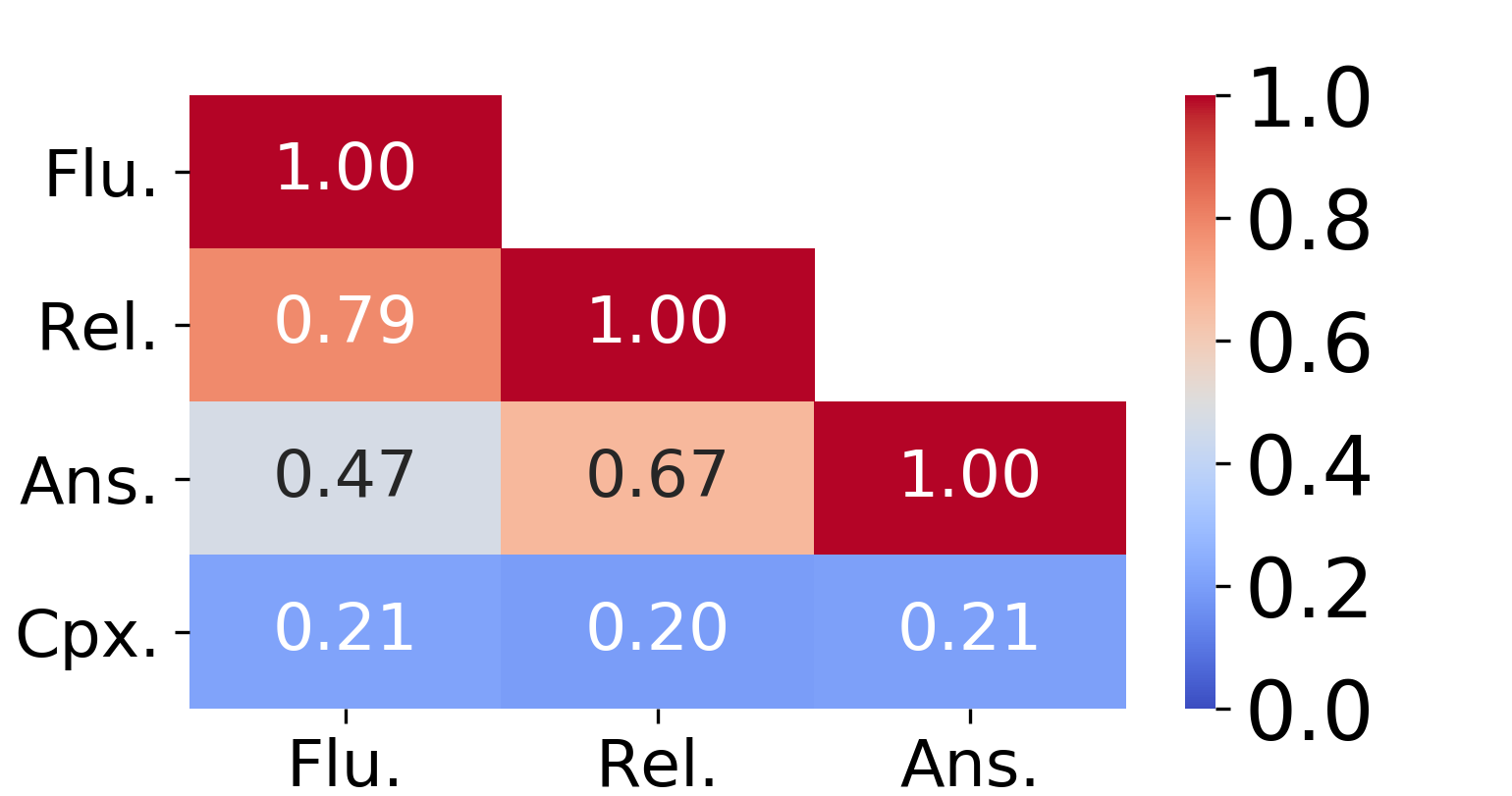}
    	\end{minipage}
    	\label{fig:corr_score}
    }
    \quad
    \subfigure[Correlation between reward scores and human ratings]
    {
    	\begin{minipage}[t]{0.4\linewidth}
    	\centering
    	\includegraphics[width=\linewidth]{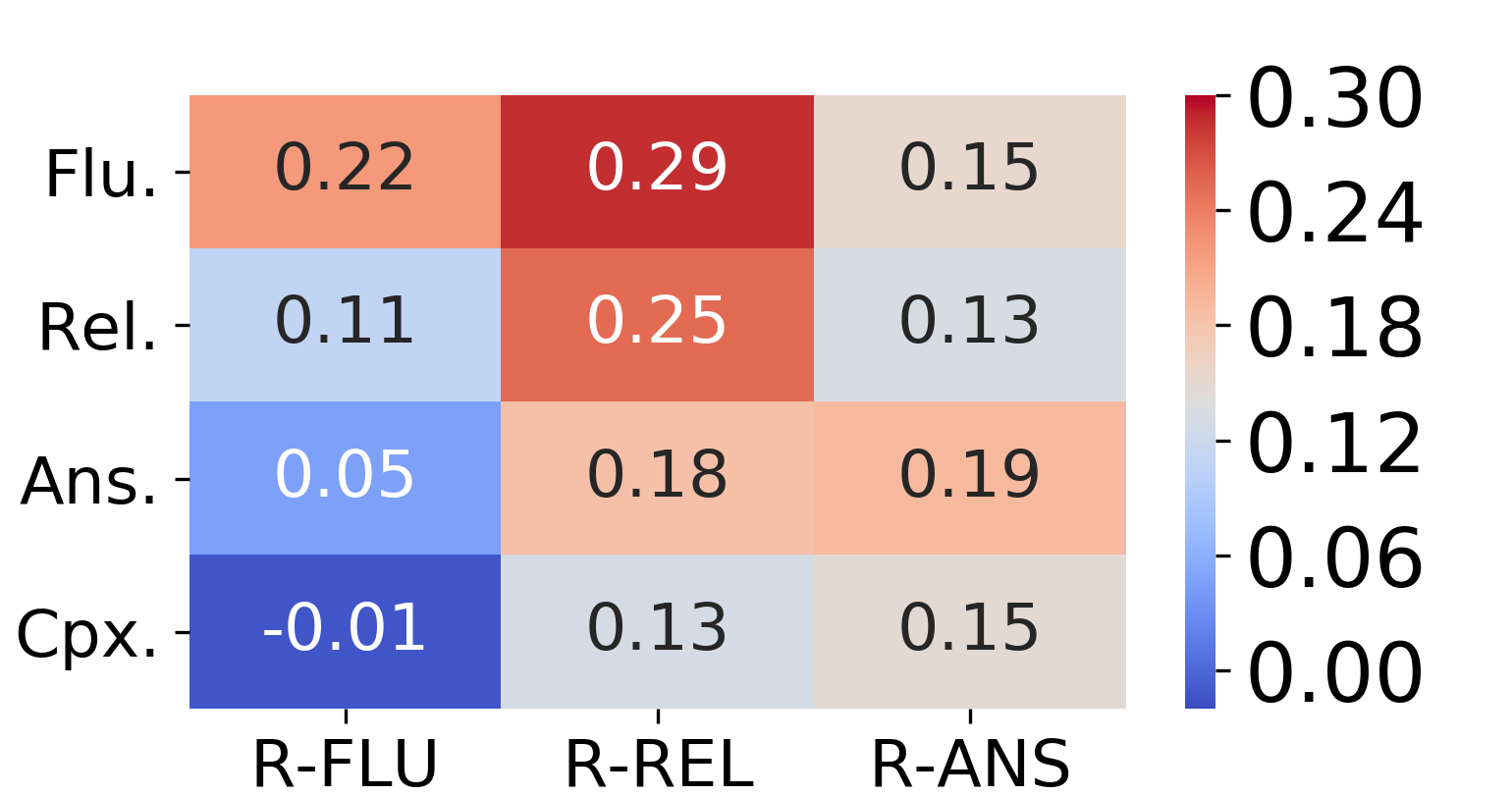}
    	\end{minipage}
    	\label{fig:corr_reward}
    }
    \caption{Heatmaps of the Pearson correlation coefficient matrices between human ratings and rewards. Flu., Rel., Ans. and Cpx. denote fluency, relevance, answerability, and complexity ratings in human evaluation, respectively. R-FLU / R-REL / R-ANS represents fluency / relevance / answerability reward.}
    \label{fig:correlation_coefficient}
\end{figure*}

We further provide a full view of how human ratings correlate to each other and how they correlate to the reward scores in Figure~\ref{fig:correlation_coefficient}. We find that the relevance rating has strong correlations with both the fluency rating ($0.79$) and the answerability rating ($0.67$), indicating that a question is more relevant to the document when it is fluent and answerable. However, a relatively weak correlation exists between the fluency and answerability, meaning a fluent question is not necessarily answerable. In Figure~\ref{fig:corr_reward}, we further find that the relevance reward has a strong correlation with not only the relevance rating, but also fluency and answerability ratings. This explains why optimizing the relevance reward alone (S2) leads to improvements on fluency and answerability as well. In contrast, R-ANS has poor correlation with Flu. and Rel., explaining why it decreases the fluency and relevance ratings. 

\subsection{Mesoscopic Analysis of Generated Questions}
\label{subsec:meso_analysis}

To further understand why the fluency and the answerability reward fail to produce a consistent judgement with humans, we conduct a mesoscopic analysis on our E4 model by comparing the generated questions receiving high rewards with those with low rewards. We detail our observations for each reward type, guided by the results in Table~\ref{tbl:meso-table}. 

% MinCR: make sure use punctuation to mark the end of headers signified by formatting. Added full stop for all headers here.
\paragraph{$\bullet$ Fluency.} From Table~\ref{tbl:meso-table} Row~F, we observe that sometimes the fluency reward is consistent with the human judgement on fluency; \textit{e.g.}, the incomplete question [FL-1] receives a low reward. However, there is often inconsistency between the fluency judged by the language model and that of human-judged fluency. For example, [FH-2] has a repetition error but is assigned a high reward, while [FL-2] with a similar repetition error receives a low reward. This is caused by the statistical bias in the language model; \textit{i.e.}, the LM tends to assign low rewards to the questions with rare or unseen entities (\textit{e.g., Kenji Mizoguchi}). The lack of commonsense knowledge is another problem of the LM: {\it e.g.}, in [FH-1] the model fails to replace the word \textit{born} with \textit{founded} to make the question logically correct. 

\paragraph{$\bullet$ Relevance.} Table~\ref{tbl:meso-table} Row R shows that the relevance discriminator judges the document--question relevance largely based on two aspects: 1) whether the question contains an entity that does not appear in the passage (ghost entity), \textit{e.g.}, \textit{Granly} in [RL-1], and 2) whether the question has a logical inconsistency with the document, \textit{e.g.}, [RL-2]. These two targets are quite consistent with the human judgement on relevance, which explains its good correlation in Figure~\ref{fig:violin-plot}. However, when the question is asking about an unmentioned aspect of an entity in the document, it is difficult for the model to assign an appropriate relevance score as in [RH-1]. A potential solution is to factor in the judgement of a good answerability discriminator (a challenge itself). 

\paragraph{$\bullet$ Answerability.} We observe in Row A that the answerability reward follows quite different criteria for whether a question can be answered compared against humans. First, most of the questions with high rewards are asking \textit{what year} (the text highlighted in pink). We find that $45.0\%$ questions generated by S3 are \textit{what year} questions, compared with $11.2\%$ for the baseline model B1. This may be caused by the data bias of the training set. Since a large portion of questions in HotpotQA are asking about date or time, this leads the QA model to learn a spurious correlation that a \textit{what year} question is more likely to be answered and hence should receive high rewards. 
% This may be caused by the incorrect bias learned by the QA model, \textit{i.e.}, it tends to regard a year in the document as an answer, which in turn causes questions asking about years to receive high answerability rewards. 
Second, when the question becomes complex, \textit{i.e.} requiring the QA model to conduct reasoning such as comparison (Questions~[AL-1] and [AL-2]) and to utilize world knowledge (\textit{e.g.} \textit{United States} is a \textit{country}), the QA model tends to give a low answerability reward. This can be explained by the insufficient ability of current QA model in answering deep questions. To improve the answerability via a QA-based reward, we believe it is crucial to address the QA model's bias in prediction and improve its reasoning ability. Otherwise, optimizing an immature QA-based reward may introduce an incorrect bias, which in turn harms the question quality. 

\begin{table}[t]
    \small
    \centering
        \begin{tabular}{c|c|l} \hline
             & \multicolumn{2}{c}{\textbf{Samples with High / Low Rewards}} \\ \hline \hline
             \multirow{4}{*}{\textbf{F.}} & \multirow{2}{*}{\textbf{H.}} & \textbf{[FH-1] Que.} Eleven : A Music Company was \colorbox[RGB]{225,197,204}{\makebox(12,3){born}} \colorbox[RGB]{250,206,248}{\makebox(40,3){in what year}} ? \\
             & & \textbf{[FH-2] Que.} Dan Smith was born \colorbox[RGB]{250,206,248}{\makebox(40,3){in what year}} ? \colorbox[RGB]{189,215,238}{\makebox(35,3){Dan Smith}} \\ \cline{2-3}
             & \multirow{3}{*}{\textbf{L.}} & \textbf{[FL-1] Que.} Park Seo - joon starred in a South Korean television series that  premiered on May 22 , 2017 \\
             & & \textbf{\textcolor{white}{[FL-1]}} every Monday and Tuesday \colorbox[RGB]{225,197,204}{\makebox(16,3){where}} ? \\
             &  & \textbf{[FL-2] Que.} Kenji Mizoguchi was born \colorbox[RGB]{250,206,248}{\makebox(40,3){in what year}} ? \colorbox[RGB]{189,215,238}{\makebox(56,3){Kenji Mizoguchi}} \\ \hline \hline
             \multirow{16}{*}{\textbf{R.}} & \multirow{9}{*}{\textbf{H.}} & \textbf{[RH-1] Doc.} " Sk8er Boi " is a song by the singer Avril Lavigne , released as the  second single from her \\
             & & \textbf{\textcolor{white}{[RH-1]}} debut album , " Let Go " ( 2002 ) . \\
             &  & \textbf{\textcolor{white}{[RH-1]} Que.} " Sk8er Boi " is a song written by a singer \colorbox[RGB]{226,208,240}{\makebox(12,3){born}} \colorbox[RGB]{250,206,248}{\makebox(40,3){in what year}} ? \\
             & & \textbf{[RH-2] Doc.} \colorbox[RGB]{211,237,226}{\makebox(38,3){Roy Holder}} then appeared in “ The Taming of the Shrew ” ( 1967 ) ,  “ Here We Go Round \\
             & & \textbf{\textcolor{white}{[RH-2]}} the Mulberry Bush ” ( 1967 ) , “ Romeo and  Juliet ” ( 1968 ) … The Taming of the Shrew is \colorbox[RGB]{211,237,226}{\makebox(3,3){a}} \\
             & & \textbf{\textcolor{white}{[RH-2]}} \colorbox[RGB]{211,237,226}{\makebox(279,3){1967 film based on the play of the same name by William Shakespeare about}} a courtship  betw- \\
             & & \textbf{\textcolor{white}{[RH-2]}} een two strong - willed people . \\
             & & \textbf{\textcolor{white}{[RH-2]} Que.} Roy Holder appeared in \colorbox[RGB]{211,237,226}{\makebox(240,3){a 1967 film based on the play of the same name by William Shake-}} \\
             & & \textbf{\textcolor{white}{[RH-2]}} \colorbox[RGB]{211,237,226}{\makebox(40,3){speare about}} what? \\ \cline{2-3}
             & \multirow{7}{*}{\textbf{L.}} & \textbf{[RL-1] Doc.} Beitun , Xinjiang is a county - level city under the  direct administration of the regional gov- \\
             & & \textbf{\textcolor{white}{[RL-1]}} ernment . \colorbox[RGB]{211,237,226}{\makebox(39,3){Wafangdian}} is one of the two northern county - level cities , the other being  Zhuanghe , \\
             & & \textbf{\textcolor{white}{[RL-1]}} under the administration of Dalian , located in the south  of Liaoning province , China . \\
             & & \textbf{\textcolor{white}{[RL-1]} Que.} Are both \colorbox[RGB]{226,208,240}{\makebox(20,3){Granly}} and \colorbox[RGB]{211,237,226}{\makebox(39,3){Wafangdian}} located in the same country ? \\
             & & \textbf{[RL-2] Doc.} In physics and engineering , the Fourier number or Fourier modulus ,  named after \colorbox[RGB]{211,237,226}{\makebox(22,3){Joseph}}  \\
             & & \textbf{\textcolor{white}{[RL-2]}} \colorbox[RGB]{211,237,226}{\makebox(24,3){Fourier}} , is a dimensionless number that  characterizes transient heat conduction . \\
             & & \textbf{\textcolor{white}{[RL-2]} Que.} \colorbox[RGB]{211,237,226}{\makebox(50,3){Joseph Fourier}} \colorbox[RGB]{226,208,240}{\makebox(54,3){was named after}} a man born \colorbox[RGB]{250,206,248}{\makebox(40,3){in what year}} ? \\ \hline \hline
             \multirow{15}{*}{\textbf{A.}} & \multirow{7}{*}{\textbf{H.}} & \textbf{[AH-1] Doc.} The Worst Journey in the World was written and published in 1922 by  a member of the exp- \\
             & & \textbf{\textcolor{white}{[AH-1]}} edition , Apsley Cherry – Garrard . \\
             & & \textbf{\textcolor{white}{[AH-1]} Que.} The Worst Journey in \colorbox[RGB]{189,215,238}{\makebox(62,3){The Worst Journey}} was \colorbox[RGB]{225,197,204}{\makebox(12,3){born}} \colorbox[RGB]{250,206,248}{\makebox(40,3){in what year}} ? \\
             & & \textbf{[AH-2] Doc.} Weber \colorbox[RGB]{211,237,226}{\makebox(25,3){' s Store}} , at 510 Main St . in Thompson Falls in Sanders  County ( founded in 1905 ) , \\
             & & \textbf{\textcolor{white}{[AH-2]}} Montana was \colorbox[RGB]{211,237,226}{\makebox(180,3){listed on the National Register of Historic Places}} in 1986 . \\
             & & \textbf{\textcolor{white}{[AH-2]} Que.} \colorbox[RGB]{250,206,248}{\makebox(40,3){In what year}} was the county founded in which \colorbox[RGB]{226,208,240}{\makebox(15,3){Terry}} \colorbox[RGB]{211,237,226}{\makebox(25,3){' s Store}} was \colorbox[RGB]{211,237,226}{\makebox(90,3){listed on the National Re-}} \\
             & & \textbf{\textcolor{white}{[AH-2]}} \colorbox[RGB]{211,237,226}{\makebox(82,3){gister of Historic Places}} ? \\ \cline{2-3}
             & \multirow{9}{*}{\textbf{L.}} & \textbf{[AL-1] Doc.} \colorbox[RGB]{211,237,226}{\makebox(179,3){John Stoltenberg is the former managing editor of}} " AARP The Magazine " , a bimonthly \\
             & & \textbf{\textcolor{white}{[AL-1]}} publication of the United States - based advocacy group AARP , a position John Stoltenberg held  \\
             & & \textbf{\textcolor{white}{[AL-1]}} from 2004 until 2012 . AARP The Magazine is an American bi - monthly magazine , \colorbox[RGB]{211,237,226}{\makebox(40,3){published by}}  \\
             & & \textbf{\textcolor{white}{[AL-1]}} the American Association of Retired People , AARP , which focuses on aging issues . \\
             & & \textbf{\textcolor{white}{[AL-1]} Que.} \colorbox[RGB]{211,237,226}{\makebox(178,3){John Stoltenberg is the former managing editor of}} a magazine \colorbox[RGB]{211,237,226}{\makebox(40,3){published by}} which organization? \\
             & & \textbf{[AL-2] Doc.} \colorbox[RGB]{211,237,226}{\makebox(50,3){8 Spruce Street}} is a 76 - story skyscraper designed by architect  Frank Gehry in the New York \\
             & & \textbf{\textcolor{white}{[AH-2]}} City . The \colorbox[RGB]{211,237,226}{\makebox(96,3){original World Trade Center}} was a large complex of seven buildings in Lower Manhattan ,  \\
             & & \textbf{\textcolor{white}{[AH-2]}} New York City , United States . \\
             & & \textbf{\textcolor{white}{[AL-2]} Que.} \colorbox[RGB]{211,237,226}{\makebox(50,3){8 Spruce Street}} and the \colorbox[RGB]{211,237,226}{\makebox(96,3){original World Trade Center}} , are located in  which country ? \\ \hline
        \end{tabular}
    \caption{Generated questions of E4, classified by high (H.) / low (L.) rewards. Colors indicate typical error types: \colorbox[RGB]{225,197,204}{\makebox(12,6){red}}: grammatical and logical errors, \colorbox[RGB]{189,215,238}{\makebox(15,6){blue}}: repetition, \colorbox[RGB]{211,237,226}{\makebox(17,6){green}}: corresponding parts between documents and questions, \colorbox[RGB]{226,208,240}{\makebox(22,6){purple}}: ghost entities and unmentioned parts, \colorbox[RGB]{250,206,248}{\makebox(15,6){pink}}: asking about \textit{year}.}
    \label{tbl:meso-table}
\end{table}

\section{Conclusion}
In this paper, we optimize three question-specific rewards via reinforcement learning on a Seq2Seq based question generator, aiming to improve the fluency, relevance and answerability of the generated questions. 
Through comprehensive analytic experiments, including automatic and human evaluation, consistency validation, and meso analysis, we show that the effectiveness of a reward is poorly reflected by automatic evaluation metrics such as BLEU. Instead, we find a reward that correlates well with the human judgement generally has better effects on improving the question quality. In future works, we believe these observations can help to guide the design of other QG-specific rewards that target on unexplored aspects of question generation, such as the informativeness and the utility of questions. 

\section*{Acknowledgements}

This research is supported in part by the National Research Foundation, Singapore under its International Research Centres in Singapore Funding Initiative, and the Natioinal Hi-Tech R\&D Program of China (No.2018YFB1005100). Any opinions, findings and conclusions or recommendations expressed in this material are those of the author(s) and do not reflect the views of National Research Foundation, Singapore. 

% include your own bib file like this:
\bibliographystyle{coling}
\bibliography{coling2020}

\end{document}